\documentclass[sigconf]{acmart}
\usepackage{graphicx}
\usepackage{caption}
\usepackage{subcaption}
\usepackage{multirow}
\usepackage{mathtools}
\usepackage{hyperref}
\usepackage[bottom]{footmisc}
\AtBeginDocument{%
  \providecommand\BibTeX{{%
    \normalfont B\kern-0.5em{\scshape i\kern-0.25em b}\kern-0.8em\TeX}}}

\setcopyright{acmcopyright}
\copyrightyear{2020}
\acmYear{2020}

\acmConference[KDD DSHealth '20]{2020 KDD workshop on Applied data science in Healthcare: Trustable and Actionable AI for Healthcare}{August 23--27, 2020}{San Diego, California, USA}
\acmPrice{15.00}
\acmISBN{978-1-4503-XXXX-X/18/06}

\begin{document}

\title{An Enhanced Text Classification to Explore Health based Indian Government Policy Tweets}


\author{Aarzoo Dhiman}
\affiliation{%
  \institution{Department of Computer Science and Engineering \\ Indian Institute of Technology Roorkee}
  \city{Roorkee, Uttarakhand-247667}
  \country{India}}
\email{aarzoo@cs.iitr.ac.in}

\author{Durga Toshniwal}
\affiliation{%
  \institution{Department of Computer Science and Engineering \\ Indian Institute of Technology Roorkee}
  \city{Roorkee, Uttarakhand-247667}
  \country{India}}
  \email{durga.toshniwal@cs.iitr.ac.in}

\renewcommand{\shortauthors}{Dhiman and Toshniwal}

\begin{abstract}
Government-sponsored policy-making and scheme generations is one of the means of protecting and promoting the social, economic, and personal development of the citizens. The evaluation of effectiveness of these schemes done by government only provide the statistical information in terms of facts and figures which do not include the in-depth knowledge of public perceptions, experiences and views on the topic. In this research work, we propose an improved text classification framework that classifies the Twitter data of different health-based government schemes. The proposed framework leverages the language representation models (LR models) BERT, ELMO, and USE. 
However, these LR models have less real-time applicability due to the scarcity of the ample annotated data. To handle this, we propose a novel\textit{ GloVe word embeddings} and class-specific sentiments based text augmentation approach (named \textit{Mod-EDA}) which boosts the performance of text classification task by increasing the size of labeled data. 
Furthermore, the trained model is leveraged to identify the level of engagement of citizens towards these policies in different communities such as middle-income and low-income groups.

\end{abstract}




\maketitle

\section{Introduction}
Indian government announces various welfare schemes for all cross-sections of the community from time to time at different levels. These schemes are primarily focused on agriculture, health, education, social welfare, energy and, e-governance, etc. 
Over the years social media data has been analyzed to perform various tasks to study the social-economic behavior \cite{huang2016activity}, linguistic characteristics \cite{preotiuc2016studying}, disaster events study \cite{kim2018emergency} and public and mental health study \cite{gkotsis2017characterisation}, etc. These analyses help in capturing the type of information shared and the groups of society influenced by such information.

 In this research work, we study the Twitter data (approx. 1.4 million tweets) from December 2018 to November 2019 pertaining to various \textit{health-related schemes}, launched by the Indian government, affecting the health status of the society directly or indirectly oriented towards various communities and societies, eg. female health, maternal health, child health, poor people health, etc. A citizen-provided feedback on the social media platforms can provide great resource for future steps to be taken by the government to develop more effective schemes and policies such that health-status of the country can be improved. 
The proposed work will help in analyzing the level of involvement of people in expressing and sharing information and opinions regarding different types of schemes. 

The government health schemes generally target the common public which comprises of the low-income and middle-income sections of the country. 
Analyzing these target groups and communities can provide deeper insights into the awareness and effectiveness of these schemes.
Thus, to cover most of these needful groups, we classify the tweets to identify the target communities/groups into \textit{general-health, poor people's health, and mother \& child health}. These target groups are further classified into respective types of engagements (\textit{informational, promotional, appreciation and complaint}). These types will help in providing the perception and level of engagement of these target groups. 
 We fine-tune the pre-trained deep learning models for language representation for text classification i.e. Bidirectional Encoder Representation from Transformers (BERT) \cite{devlin2018bert}, Embeddings from Language Models (ELMO) \cite{peters2018deep} and Universal Sentence Encoder (USE) \cite{cer2018universal} to generate highly accurate results. However, the application of these powerful models on real-time problems (or real-time data) requires an ample amount of annotated data for supervised learning. The process of manual annotation is a laborious and time-intensive task, which restrains the application of these models on the real-time dataset. Thus, this research work proposes to integrate a novel text augmentation (\textit{Mod-EDA}) with the underlying classification model. Mod-EDA proposes to increase the size of the annotated data set while overcoming the limitations of previous models. Mod-EDA handles this by fine-tuning GloVe generated domain-specific word embeddings to perform class-wise insertions and substitutions in the text. We use the word embeddings to generate a set of similar words from the corpus. However, word embeddings tend to generate similar words based on contexts and lose sentiments. Mod-EDA overcomes this problem by maintaining the text level and tweet level sentiment polarity while augmenting the text. Further details can be found in the later sections. 

\section{Text Classification Framework}

\subsection{Dataset Description: Data Collection and Preprocessing}
There is a wide range of government schemes and policies launched for the generation of employment, poverty alleviation, providing education and public health facilities from time to time. 
A list of 106 government schemes launched in India, which are related directly or indirectly to health, are used as the seed/hashtags to collect the tweets. A sample hashtags used are \#IPledgeFor9, \#MaternalHealth, \#NHPIndia, JananiSurakshaYojna, Ujjwalayojna, PoshanAbhiyan, IndiaMGNREGA, MGNREGA, Sukanya Samriddhi Yojna, betibachaobetipadhao, Deen Dayal Antyodaya Yojana, Rashtriya Mahila Kosh etc. 
The data collection was done using the Twitter Streaming API between December 2018 to November 2019, which collected the tweets containing one or more of these hashtags resulting in 3.2 million tweets. 


The raw Twitter data collected using the Streaming API consists of a wide range of attributes e.g., timestamp, text, user details, hashtags, place, URLs, and user mentions, etc. However, for the current research work, only a few attributes are considered useful. Thus, the tweets and retweets were parsed to extract the values to attributes such as timestamp, text, place, and user name from the tweets.
The tweets were cleaned to remove all the stop-words, hyperlinks, special symbols, recurring characters, and other non-decodable information. The dataset contained the keywords which were very general, such as \textit{`MaternalHealth'}, and \textit{`MotherCare'} etc. Thus, to keep the tweets relevant to the current study, the tweets posted from outside India were removed. The location information was extracted by geoparsing the user-location from the tweets using the python module \textit{geopy}. The tweets having a language other than English were also removed from the set of relevant tweets.

\subsection{Two-level Tweet Classification Framework}

\subsubsection{Expert Labeling}
\label{sec:label}

Due to the polysemy of the hashtags, some of the tweets may found to be irrelevant for the study. For example, the hashtag ``NITI AAYOG" (a government scheme) may extract tweets that mention 'Niti' as the name of a person.  However, it is difficult to handle such pragmatic ambiguities using keywords based approach; thus, supervised text classification is used to remove irrelevant tweets. First a sample of 4000
tweets were labeled into two categories, namely, Relevant and Irrelevant ensuring 2000 tweets in each class. All the tagging was done by three Master's students and the label having a $2/3^{rd}$ probability of being chosen was taken to be the final label. We trained three text classification models i.e. BERT, ELMo and USE on these 4000 tweets. The accuracy for these models were found to be 0.92, 0.90 and 0.91 respectively. Next, we used the BERT trained classification model to predict the relevance of tweets. As a result, approx. 1.4 million tweets remained as the \textit{relevant tweets} and the rest of the tweets were discarded. The relevant 2000 tweets were later labeled into four categories in terms of focus/target-group of the scheme i.e. General Health-related, Poor people's health-related, Maternal health-related, and Election-related tweets. Next, to make the study more informative, the tweets were further divided into four categories i.e., \textit{Informational tweets, Promotional tweets, Appreciation tweets, and Complaint tweets}. The fundamental difference between Informational and Promotional tweets is that informational tweets are more oriented towards spreading information and the promotional tweets tend to promote the after-effects or outcomes (generally positive) of these schemes. The election-related category consists of the group of tweets that mentioned the schemes and the election information at the same time. These tweets were segregated to keep such election campaign-related tweets separate from the pure schemes related tweets. 


\subsubsection{Mod-EDA}

Easy Data Augmentation (EDA) \cite{wei2019eda} is a text augmentation method that can generate approximately nine times the original data with only simple modifications i.e., Random Insertion, Synonym Replacement, Random Deletion, and Random Swap. EDA performs a random combination of one to all types of augmentations on a sentence to generate the new dataset. Mod-EDA modifies the Random insertion and Synonym Replacement to more domain-specific, sentiment reactive, and contextual Insertion and Substitution operations, respectively. Mod-EDA is based on the premises that
``A sentence should be augmented such that it contains the random domain-specific knowledge while maintaining the semantic meaning of the sentence." The \textit{randomness} helps to add the required noise or variance in the data to maintain the data distribution, while the domain and context-sensitivity tends to keep the augmentation problem-specific. 
 The random deletion and random swapping has been done similar to that of EDA. The following paragraphs explain the limitations in EDA and proposed modifications in detail. 

\begin{itemize}
\item \textit{Modified Substitution}: EDA chooses a word randomly and replaces it with the synonym. However, replacing a word with a synonym may not always result in the domain-specific keyword. The domain of the data decides the context of the word. For example, \textit{Mosque} is considered a holy place and is expected to have a positive sentiment in the sentence. However, if the data belongs to some attack on the Mosque, the sentiments of sentences having word \textit{Mosque} change to negative. Thus, maintaining the contextual information and keeping substitutions domain-specific is important for text augmentation. Similarly, replacing a word \textit{'back'} with \textit{'backwards'} does not add any randomness to the augmented data. 
We first extract a set of $t$ similar words using the GloVe generated word embeddings to be used as candidate for substitution.
However, word embeddings generate similar words that may not be of similar sentiment. For example, the similar words for \textit{great} are \{\textit{well, good, minor, little}.. \}. The reason behind\textit{ minor, little} belonging to the similar word list of \textit{great} is that all these words are generally used in similar contexts. Thus, to maintain the semantic meaning of the sentence, the word is substituted with a word having the most closest sentiment. The sentiment are calculated using AFINN\footnote{\url{https://github.com/fnielsen/afinn}} lexicon polarity dictionary.
 
\item \textit{Modified Insertion}: EDA chooses an index randomly and inserts a random word from the corpus at that index to perform the random insertion. However, random insertion may insert words that are \textit{negations} e.g. not, never etc. (inverting the polarity), \textit{intensifiers} e.g. very, much etc. (increasing or decreasing the polarity of the sentence) and \textit{modals} e.g. must, can etc. (introducing the context of possibility or necessity) etc. These insertions fail to maintain the accurate sentiments of the sentence. 
Thus, to keep the sentiment of the sentence intact, we generate a list of class-specific frequent keywords. A random selection of $t$ words is extracted from the frequent keywords list, and a list of similar words to that $t$ words is generated using the GloVe word embeddings. The word closest to the polarity of the original sentence is chosen to be inserted into the sentence.

\end{itemize}

\section{Results and Discussions}

\subsection{Tweet Classification}

The performance of the proposed Mod-EDA is tested on the text classification task with BERT, ELMO, and USE. The results are an average of five random seeds. Before applying Mod-EDA on Twitter dataset, the accuracy of Mod-EDA is compared with EDA on a benchmark dataset SST2 (Stanford SentiTreebank dataset 2013) \cite{socher2013parsing}. We run BERT, ELMO, and USE on the dataset without any augmentation, with EDA, and with Mod-EDA. 
The accuracy of all the models were captured in terms of precision, recall, and F1-score using 10-fold cross-validation. For testing the accuracy of the classification model, 90 percent of the data is separated from the whole corpus, and Mod-EDA and EDA are applied on it to generate an augmented corpus. The training and testing set is generated from this 90 percent data in the 80-20 ratio. This is named as the validation accuracy of the data. Later the trained model is run on the remaining 10 percent of the data. The accuracy of the remaining 10 percent data is named as the testing accuracy. This type of training-testing ensures that there is no overlap between the training and the testing data. 
The average performance is shown in Table \ref{tab:acc_comp0}. Mod-EDA shows an average improvement of 1.76\% from EDA (maximum 2.03\% with BERT) and 1.82\% (maximum 2.98\% with BERT) from data without augmentation in F1-score is visible.

\begin{table}[!t]
\centering
\caption{Comparison of Mod-EDA, EDA and without augmentation on SST2 dataset.}
\label{tab:acc_comp0}
\resizebox{\columnwidth}{!}{
\begin{tabular}{|l|l|l|l|l|l|l|}
\hline
\multirow{2}{*}{} & \multicolumn{3}{c|}{Validation Results} & \multicolumn{3}{c|}{Testing Results}  \\ \cline{2-7} 
                  & Precision    & Recall      & F1 Score   & Precision & Recall                        & F1 Score                      \\ \hline
BERT              & 0.7030    & 0.7498  & 0.749855   & 0.7153  & \multicolumn{1}{l|}{0.7704} & \multicolumn{1}{l|}{0.7418} \\ 
EDA+BERT          & 0.9075     & 0.9342    &\textbf{ 0.9207}    & 0.7201  & \multicolumn{1}{c|}{0.7589} & \multicolumn{1}{c|}{0.7390} \\ 
Mod-EDA+BERT      & 0.9052     & 0.9257  & 0.9153   & 0.7414  & \multicolumn{1}{l|}{0.76771}  & \multicolumn{1}{l|}{\textbf{0.7543}}  \\ 
ELMO              & 0.7130     & 0.7298   & 0.7213   & 0.6824  & 0.7225                     & 0.7018                       \\ 
EDA+ELMO          & 0.9043     & 0.9234   & \textbf{0.9137}    & 0.7028  & 0.7223                       & 0.71225                      \\
Mod-EDA + ELMO    & 0.8924      & 0.92543     & 0.9086    & 0.7134  & \multicolumn{1}{l|}{0.7323} & \multicolumn{1}{l|}{\textbf{0.7227}}  \\ 
USE               & 0.7301     & 0.7198    & 0.7252   & 0.6924  & 0.7223                       & 0.7070                      \\ 
EDA+USE           & 0.8923      & 0.9123      & \textbf{0.9022}     & 0.6902   & \multicolumn{1}{l|}{0.7112} & \multicolumn{1}{l|}{0.7005} \\ 
Mod-EDA+USE       & 0.8935      & 0.8933     & 0.8934    & 0.7034   & 0.7223                      & \textbf{0.7127}                     \\ \hline
\end{tabular}}
\end{table}

The tweets were classified into multiple categories on multiple levels, as explained above. Table \ref{tab:acc_comp} summarizes the recall, precision, F1-score, and accuracy of three state-of-the-art pre-trained language representation models BERT, ELMO, and USE with Mod-EDA. The accuracy of classification was found to be best with BERT. The fine-tuned BERT model was further applied to the full Twitter corpus for the duration of Dec 2018 to Nov 2019 to predict the different categories of the tweets. 

\begin{figure}[!t]
\includegraphics[width=6cm]{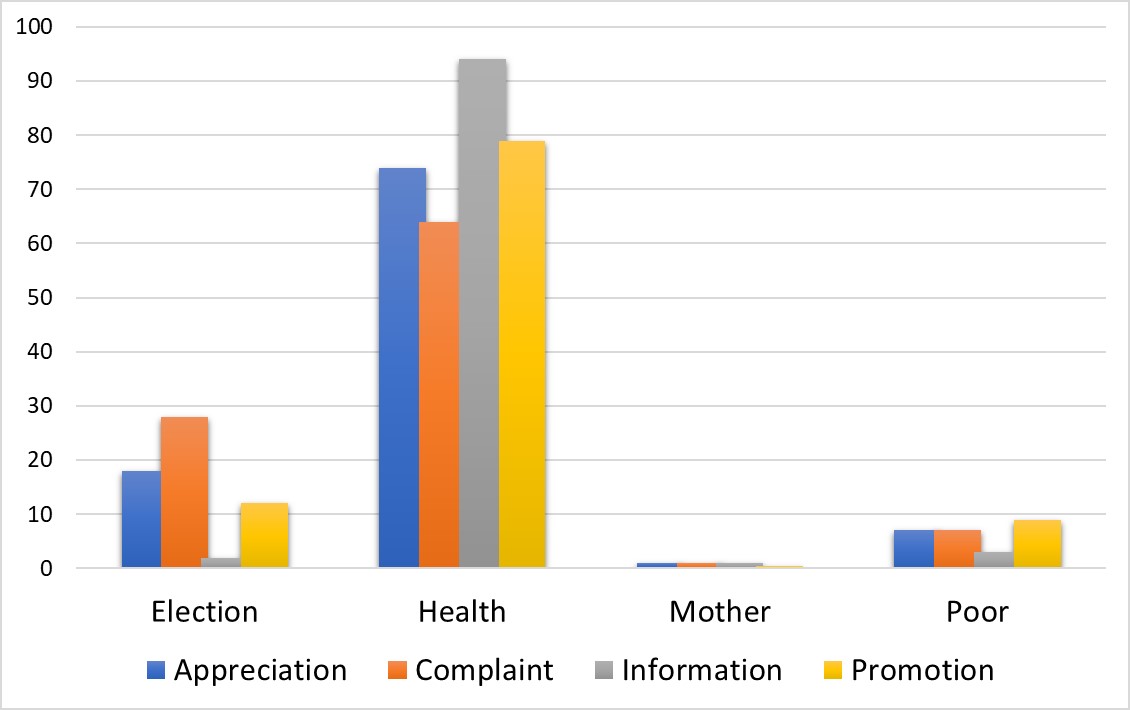}
\caption{The bar plot representation of percentage number of tweets belonging to different types in different domains}
\label{fig:multibar} 
\end{figure}

Figure \ref{fig:multibar} provides the bar plot representation of the percentage of tweets in different categories. The figure depicts that the maximum proportion of tweets belong to general health-related schemes and the minimum proportion of tweets belong to the maternal health-related schemes. The high frequency of health-related schemes is obvious due to a bigger target domain and the success and popularity of schemes targeted to the common public, i.e., health insurance schemes such as Ayushman Yojana (PMJAY)\footnote{\url{https://www.pmjay.gov.in}}. 
An approximately 40\% increase in the internet user population with 25\% internet penetration in rural areas of the country also adds up to this result.

\begin{table}[!b]
\centering
\caption{Accuracy measurement of BERT, ELMO and USE for fine tuning the labeled dataset}
\label{tab:acc_comp}
\resizebox{\columnwidth}{!}{
\begin{tabular}{|l|l|l|l|l|l|l|l|l|l|l|}
\hline
\multicolumn{2}{|l|}{\multirow{2}{*}{}}                                                                                                                   & \multicolumn{3}{c|}{Mod-EDA +BERT} & \multicolumn{3}{c|}{Mod-EDA+ELMO}                                & \multicolumn{3}{c|}{Mod-EDA+ USE} \\ \cline{3-11} 
\multicolumn{2}{|l|}{}                                                                                                                                    & Precision   & Recall   & F1-Score  & Precision                 & Recall                    & F1-Score & Precision   & Recall  & F1-Score  \\ \hline
\multirow[t]{5}{*}{\begin{tabular}[t]{@{}l@{}}Target Groups/\\ Communities\end{tabular}} & Election                                                          & 0.71        & 0.71     & 0.71      & \multicolumn{1}{l|}{0.75} & \multicolumn{1}{l|}{0.64} & 0.69     & 0.67        & 0.57    & 0.62      \\ 
                                                                                      & \begin{tabular}[c]{@{}l@{}}General-\\ health related\end{tabular} & 0.9         & 0.93     & 0.92      & \multicolumn{1}{l|}{0.85} & \multicolumn{1}{l|}{0.95} & 0.89     & 0.81        & 0.93    & 0.86      \\ 
                                                                                      & \begin{tabular}[c]{@{}l@{}}Poor people's \\ health\end{tabular}   & 0.9         & 1        & 0.95      & \multicolumn{1}{l|}{0.69} & \multicolumn{1}{l|}{1}    & 0.82     & 0.82        & 1       & 0.9       \\ 
                                                                                      & \begin{tabular}[c]{@{}l@{}}Mother-child \\ health\end{tabular}    & 0.78        & 0.7      & 0.74      & 0.92                      & 0.6                       & 0.73     & 0.71        & 0.5     & 0.59      \\ 
                                                                                      & Accuracy                                                          &             &          & \textbf{0.8557 }   &                           &                           & 0.8269   &             &         & 0.7789    \\ 
\multirow[t]{5}{*}{\begin{tabular}[t]{@{}l@{}}Type of \\ Engagement\end{tabular}}        & Appreciation                                                      & 0.66        & 0.57     & 0.61      & \multicolumn{1}{l|}{0.65} & \multicolumn{1}{l|}{0.67} & 0.66     & 0.44        & 0.5     & 0.47      \\ 
                                                                                      & Complaint                                                         & 0.65        & 0.85     & 0.73      & 0.62                      & 0.77                      & 0.69     & 0.64        & 0.69    & 0.67      \\
                                                                                      & Information                                                       & 0.65        & 0.63     & 0.64      & \multicolumn{1}{l|}{0.44} & \multicolumn{1}{l|}{0.47} & 0.46     & 0.39        & 0.53    & 0.45      \\ 
                                                                                      & Promotion                                                         & 0.6         & 0.63     & 0.61      & 0.55                      & 0.56                      & 0.55     & 0.7         & 0.53    & 0.61      \\ 
                                                                                      & Accuracy                                                          &             &          &\textbf{ 0.697}     &                           &                           & 0.6432   &             &         & 0.5537    \\ \hline

\end{tabular}}
\end{table}

\begin{figure}[!b]
\centering
\begin{minipage}[b]{0.5\columnwidth}
\includegraphics[width=4.5cm]{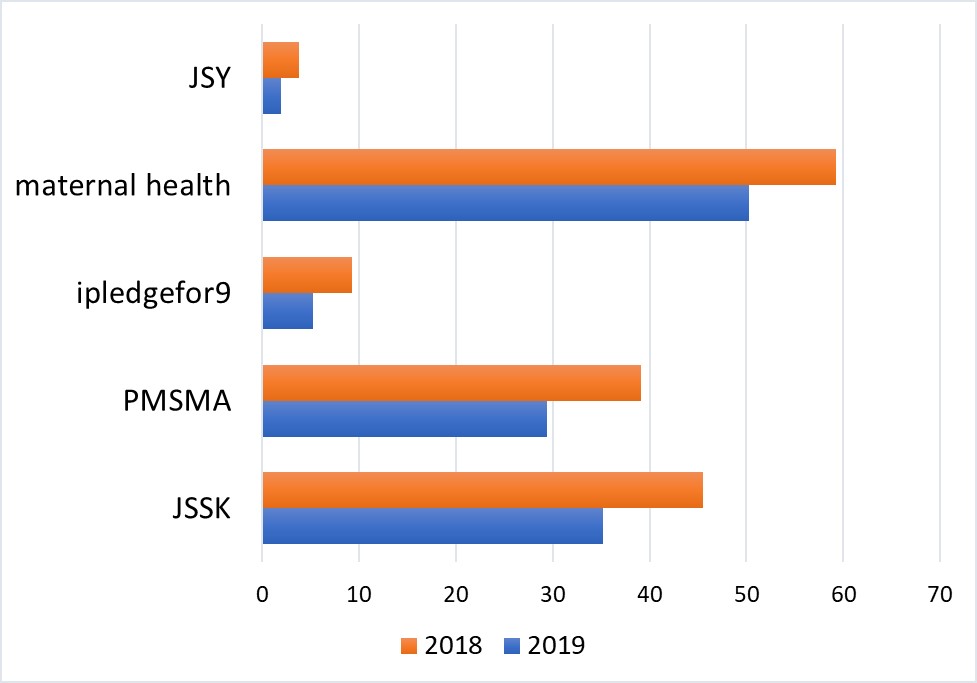}\\
\subcaption{Maternal health related hashtags}
 \label{fig:mother}
\end{minipage}%
\begin{minipage}[b]{0.5\columnwidth}
\includegraphics[width=4.5cm]{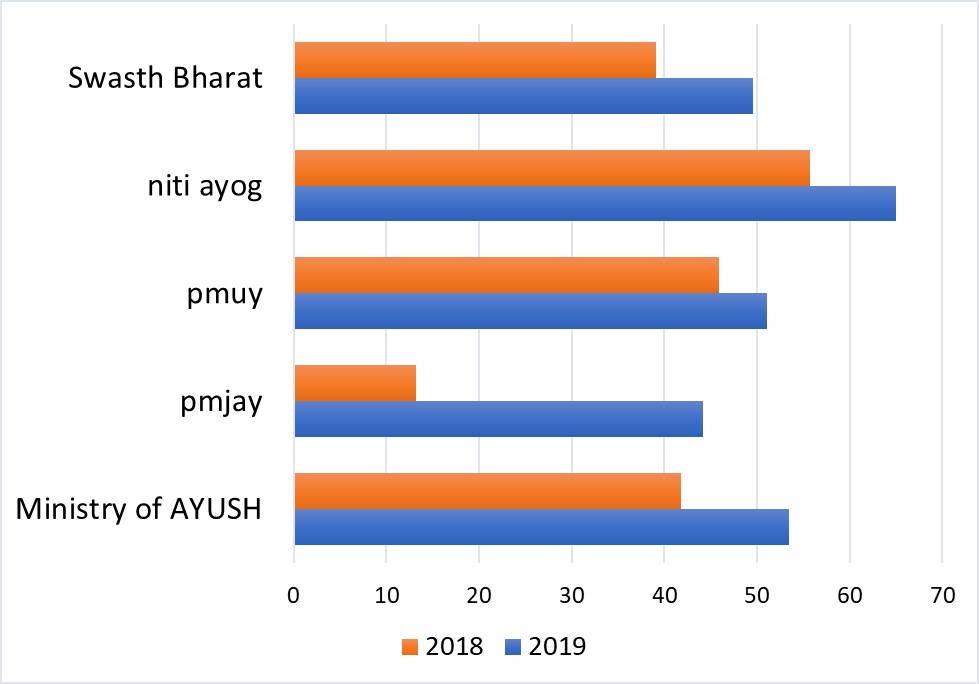}\\
\subcaption{Poor people health related hashtags}
\label{fig:poor}
\end{minipage}%
\caption{The number of GT queries related to mother-child healthcare scheme and poor people health schemes related popular hashtags}
 \label{fig:gt_plot} 
 \end{figure}

In a study provided by World Health Organization\footnote{\url{https://www.who.int/southeastasia/news/detail/10-06-2018-india-has-achieved-groundbreaking-success-in-reducing-maternal-mortality}} India achieved the groundbreaking success in lowering the maternal mortality ratio (MMR)\footnote{\url{https://data.unicef.org/topic/maternal-health/maternal-mortality/}} by 77\% in India in 2018. According to Unicef/WHO, one of the primary reasons behind this success was introduction of government schemes such as the Janani Shishu Suraksha Karyakram\footnote{\url{https://www.nhp.gov.in/janani-shishu-suraksha-karyakaram-jssk_pg}}(JSSK) which provides free transport services, no expense delivery and supervised rural births and the Pradhan Mantri Surakshit Matritva Abhiyan\footnote{\url{https://pmsma.nhp.gov.in}} (PMSMA) providing women access to antenatal check-ups, obstetric gynecologists and tracking high pregnancies, etc. This emphasizes the fact that the introduction of maternal health-related schemes has played a major role in improving maternal health standards of women in India. However, this is news of 2018, and the data collection belongs to the year 2019. Thus, social media and the government might have shifted its focus from mother-child healthcare schemes and moved towards schemes focusing on other groups of the society in 2019. Figure \ref{fig:multibar} depicts that the focus may have shifted to poverty-stricken citizens because it has the highest frequency of tweets after the general health-related schemes. To validate this hypothesis, we use the Google Trends (GT) data of the five most frequent maternal health and poor people health schemes related hashtags and keywords for the year 2018 and 2019. The GT data is used because we did not have the Twitter data available for 2018. The GT data is relative to the highest number of queries in the given duration. Thus, we take the average of the queries for the whole year to create a bar plot representation for these frequent keywords, as given in Figure \ref{fig:gt_plot}. The graph depicts that there is a decrease in the number of maternal health schemes related queries and an increase in queries concerning poverty-stricken public health schemes from 2018 to 2019. This validates the hypothesis that the government tends to shift the focus from one target domain to another domain with time and success in previous targets.

Furthermore, the data is specific to the government generated schemes and campaigns, and most of the tweets are informational and promotional, as depicted in Figure \ref{fig:multibar}. 
As per the statistics, India has a 65.97\% rural-area based population, and in 2018, an increase of 35\% (from 9\% in 2009) in the internet users in rural areas of the country was reported stating that ``an excess of half a billion internet users is driven by the rural internet growth and usage". \footnote{\url{https://economictimes.indiatimes.com/tech/internet/india-has-second-highest-number-of-internet-users-after-china-report/articleshow/71311705.cms?from=mdr}} As a result, the awareness of health-care schemes targeting poor people has increased over time. The government is also putting closer attention on the health-based schemes which target the health of the poverty-stricken public. Thus, it can be considered an important initiative by the government to spread the information that is useful for the population-in-need by providing social media coverage. A regular dissemination of such information motivates the everyday users to provide their views and perceptions on such content and share and spread their detailed knowledge. The high amount of Complaint related tweets shows that people also tend to share the negative news along with the hashtags of these schemes. Also, the Appreciation related tweets are the least occurring of all. This is also a general phenomenon as per the previous studies that tells that there are very few tweets that provide any sentiments and reactions towards these schemes.

\section{Conclusion}

The proposed framework overcomes the limitation of deep learning based classification models i.e., scarcity of annotated data for supervised learning. Mod-EDA, proposed as an extension of EDA, leverages the dense GloVe based word vector representations and the class-wise sentiments to make the augmented sentences more context and domain sensitive. Mod-EDA is tested on state-of-the-art language representation models i.e., BERT, ELMO, and USE, which attains approx 2-3\% of improvement in accuracy of the text classification task. 
Furthermore, the analysis on government scheme related Twitter data helped us conclude that the government tends to shift its focus from domain to domain after achieving success. This theory was validated by the Google Trends data from 2018 and 2019.  
For future work, the correlation between the Indian general election results and government schemes Twitter data can be identified.


\bibliographystyle{ACM-Reference-Format}
\bibliography{sample-base}


\end{document}